\documentclass{article}

\usepackage[preprint]{corl_2021}
\usepackage[pdftex]{graphicx}
\usepackage{subcaption}
\usepackage{amsmath}
\usepackage{amssymb}
\usepackage{threeparttable}
\usepackage{booktabs} % for better looking tables
\usepackage{siunitx}  % for units of measure and data in tables
\usepackage{algorithmic}
\usepackage[ruled,vlined]{algorithm2e}

\usepackage{xcolor}
\DeclareMathOperator*{\argmax}{arg\,max}

\title{%Interferobot v2
Aligning an optical interferometer with beam divergence control and continuous action space}

\usepackage{authblk}

\author[1,2,*]{\textbf{Stepan Makarenko}}
\author[1]{\textbf{Dmitry Sorokin}}
\author[1]{\textbf{Alexander Ulanov}}
\author[1,3]{\textbf{A. I. Lvovsky}}
\affil[1]{Russian Quantum Center, Moscow, Russia}
\affil[2]{Moscow Institute of Physics and Technology, Russia}
\affil[3]{University of Oxford, United Kingdom}
\affil[*]{makarenko.sd@phystech.edu}

\begin{document}
\maketitle

%===============================================================================
% Transfer of reinforcement learning agents from simulated environments to the real-world will have a lot of practical applications
% In this work we consider task of vision-based alignment of an optical Mach-Zehnder interferometer with continuous action space and new optical elements
% Continuous actions agent trains only in a simulated environment with enhanced domain randomizations
% In experimental evaluation continuous actions agent significantly outperforms the discrete one and the human expert which makes it superhuman in the interferometer alignment.
\begin{abstract}
    Reinforcement learning is finding its way to real-world problem application, transferring from simulated environments to physical setups. In this work, we implement vision-based alignment of an optical Mach-Zehnder interferometer with a confocal telescope in one arm, which controls the diameter and divergence of the corresponding beam.  We use a continuous action space; exponential scaling enables us to handle actions within a range of over two orders of magnitude. Our agent trains only in a simulated environment with domain randomizations. In an experimental evaluation, the agent significantly outperforms an existing solution and a human expert.

    %%Reinforcement learning is finding its way to real-world problem application, transferring from simulated environments to physical setups. In this paper, we extend our previous approach for vision-based alignment Mach-Zehnder interferometer to continuous action space and new optical elements. This work uses logarithmic scaling to handle actions with different orders of amplitudes. This work also enhances previous domain randomizations and being trained only in simulation our new agent significantly outperforms the previous one, and human-expert making it superhuman in the considered task.
\end{abstract}


%===============================================================================

\section{Introduction}

Reinforcement learning (RL) demonstrates incredible success in simulated environments, surpassing a human expert in Atari \citep{mnih2013playing}, chess, shogi and Go \citep{silver2018general}, as well as in more complicated games such as Dota2 \citep{berner2019dota} and StarCraft \citep{vinyals2019grandmaster}. In robotics, reinforcement learning shows remarkable results in pushing \citep{peng2018sim}, grasping \citep{kalashnikov2018scalable} and stacking objects \citep{haarnoja2018composable}. RL enables robots to walk, overcoming different obstacles \citep{haarnoja2018learning} and learn agile animal skills \citep{peng2020learning}.  However, RL agents are not yet widely used in real robotics and can hardly be compared to humans in a physical environment.

The principal problems of real-world applications of RL are nonstationarity and stochasticity of the physical environment, complexity and time intensiveness of data acquisition as well as unsafety of training and evaluation of policies on real robots \citep{dulac2019challenges}. A common method to handle these challenges is to train an agent in a high-fidelity simulation (source domain) of the real environment (target domain). An agent can then  be transferred to a physical robot. The main limitation of this approach is inevitable discrepancy between the source and target domains that lead to performance loss after transfer. This limitation can be mitigated with the randomization of simulator parameters \cite{peng2018sim, peng2020learning, tobin2017domain}. 

A promising application domain for smart robotics is experimental optics. Optical physicists work with installations consisting of hundreds or thousands of elements. These setups need to be thoroughly aligned before measurements can be conducted. Dependent on the setup size, such a setup can require several hours of work by a group of experts on a daily basis. The automation of this procedure could drastically enhance the productivity of research groups. Since the alignment is an iterative decision-making process with a well-defined reward function, it is natural to formulate and solve it in terms of reinforcement learning.

The alignment of a large optical setup can typically be separated into modular tasks which can be completed in sequence. A common modular task is the alignment of a Mach-Zehnder interferometer (MZI), which consists in matching the optical modes of the electromagnetic waves in two paths so that they exhibit a visibility of unity when interfering with each other. The robotization of MZI alignment was studied by Sorokin {\it et al.} \citep{sorokin2020interferobot}, where this problem was treated as a partially observable Markov decision process, with the observation being a sequence of interference patterns observed on a camera as the relative phase of the two arms is being varied. They trained a discrete-action (VD D3QN \citep{huang2018vd}) agent, dubbed \emph{Interferobot}, in a simulated environment with domain randomizations and successfully transferred the trained agent to a real setup. The trained agent achieved superhuman-level performance in terms of speed and quality of alignment. 

However, the task of Ref.~\citep{sorokin2020interferobot} was limited to matching the spatial positions and directions of the two otherwise identical beams. It is much simpler than those alignment tasks that occur in practice, which additionally require  matching the geometric sizes of the beams, their divergence, and, in the case of pulsed lasers, arrival time. These additional features of the experiment result in more complex interference patterns (a richer observation space in RL parlance), hence requiring a more sophisticated agent for their interpretation. %The original Interferobot suffers from a number of limitations that set these tasks off limits to this agent. First, the number of discrete actions grows exponentially with the number of degrees of freedom \citep{dulac2015deep}. In Ref.~\citep{sorokin2020interferobot}, this issue was circumvented by selecting and operating only one of the controls at each moment in time; however, such an approach significantly lengthens the alignment procedure. Second, the actions associated with each control span multiple orders of magnitude: rough adjustments in the beginning, followed by finer and finer steps as the alignment progresses. Implementing this in a discrete action space complicates the heuristic handpicking of the appropriate action sizes and greatly enlarges the action space. Third, the interference patterns obtained from beams with different sizes and divergences are more difficult to interpret than those exhibited by identical beams, so an agent would require a more complex \textcolor{red}{policy}. 

\textbf{The main contribution of our work} is to develop a new agent for interferometer alignment, which addresses this challenge. Our solution is based on a different RL algorithm and features a number of important innovations. \emph{First,} the action space was changed from discrete to continuous, which also changed the policy task from classification to regression. This enabled us to solve the issue arising in the discrete action space, consisting in  the number of actions growing exponentially with the number of degrees of freedom \citep{dulac2015deep}. However, it raises a new concern: the agent's actions towards the end of the alignment process need to be much finer than in the beginning; the action space spans over two orders of magnitude. We address this with our \emph{second} innovation --- exponential action rescaling,  which allows the agent to effectively explore actions of different magnitudes. \emph{Third,} by granting the agent a continuous action space, we run a risk to leave the safe space of experimental parameters, which may even result in damaging the equipment. We solve this issue by introducing penalties that discourage the policy from approaching boundaries. \emph{Fourth,} we enhance the set of domain randomizations that help our agent perform better in the sim-to-real transfer.  Our resulting agent\footnote{\url{https://github.com/Stepan-Makarenko/RL_interferometer_alignment}} significantly outperforms the original Interferobot and a human expert in terms of both time and quality of the alignment.

%1) We add a system of two optical lenses in one of the interferometer arms, which leads to richer state and observation spaces that represent real optical experiments more closely. 2) We demonstrate that an optical interferometer can be successfully aligned by an agent with a continuous action space. The agent trained fully in a simulated environment can easily leverage multidimensional actions on a real setup. 3) We use exponential action rescaling which allows the agent to effectively explore actions of different magnitudes. 4) We enhance the set of domain randomizations that let our agent perform better in the sim-to-real transfer.  

%===============================================================================

\section{Related works}
RL agents for robotics can be trained either on real-world data or in simulation. The advantage of the former is that the agent receives hands-on experience of the environment in which it will be tested. The shortcoming is that the acquisition of a large dataset required for training  is a complicated and time-consuming process. This is especially the case when the observations are vision-based. For example, the training of  a grasping agent in \citet{kalashnikov2018scalable} required several weeks, in spite of parallelization across 7 identical robots. This complication can be addressed with the help of behavior cloning: \citet{vecerik2019practical} trained an agent solving the insertion task using raw visual images using dozens of expert successful and failed trajectories. Hands-on training is simplified for agents without visual observations. For example, \citet{haarnoja2018learning} developed  a quadruped walking robot that can generalize to unseen terrains with the training requiring as little as two hours. Another approach for training robot locomotion was presented by \citet{yang2020data}, who collected 4.5 minutes worth of data from a simple quadrupedal robot to model the robot's dynamics and used this model to real-time action planning. 

An alternative approach is to train the agent in simulation and transfer it to a real-world system afterwards. To eliminate the discrepancy between the real and simulated environments, \citet{tobin2017domain} proposed a simple technique called domain randomization: they trained an object detection model on simulated images with different textures, lights, object and camera positions, and demonstrated that such a model achieves high accuracy in the real world. \citet{peng2018sim} trained a robotic hand to push an object using randomization of physical parameters such as friction, mass, damping, etc. Another example is the vision-based task of dexterous in-hand manipulation \citep{andrychowicz2020learning}, which demonstrated the effectiveness of applying randomizations to both the physical parameters and image observations.

% Another vision-based task of dexterous in-hand manipulation \citep{andrychowicz2020learning} was solved using both randomizations of physical parameters and image observations.

% Переписать
The automation of optical system alignment is constantly evolving. In 1987 \citet{gabler1987optical} demonstrated automated alignment of an optical fibre used an iterative algorithm that consequently finds the maximum of a photo-detector signal with respect to each of the three fibre movement axis.
\citet{fang2016automated} considered automated alignment of a system of two lenses with 8 degrees of freedom. Acquiring a focal plane image, their algorithm performs principal component analysis and Kalman filtering to calculate proper control inputs.

Deep machine learning, particularly RL, has become increasingly popular in experimental optics. For example, \citet{sun2020deep} used an RL agent to  stabilize a mode-locked
laser by controlling waveplates and polarizers. RL algorithms are also  routinely used to optimize optical communications \cite{wang2021artificial}, e.g.~to route traffic in optical transport networks \citep{suarez2019routing,li2020digital}.

%===============================================================================

\section{Mach-Zehnder interferometer}
 
\begin{figure}[ht]
  \includegraphics[width=0.8\linewidth]{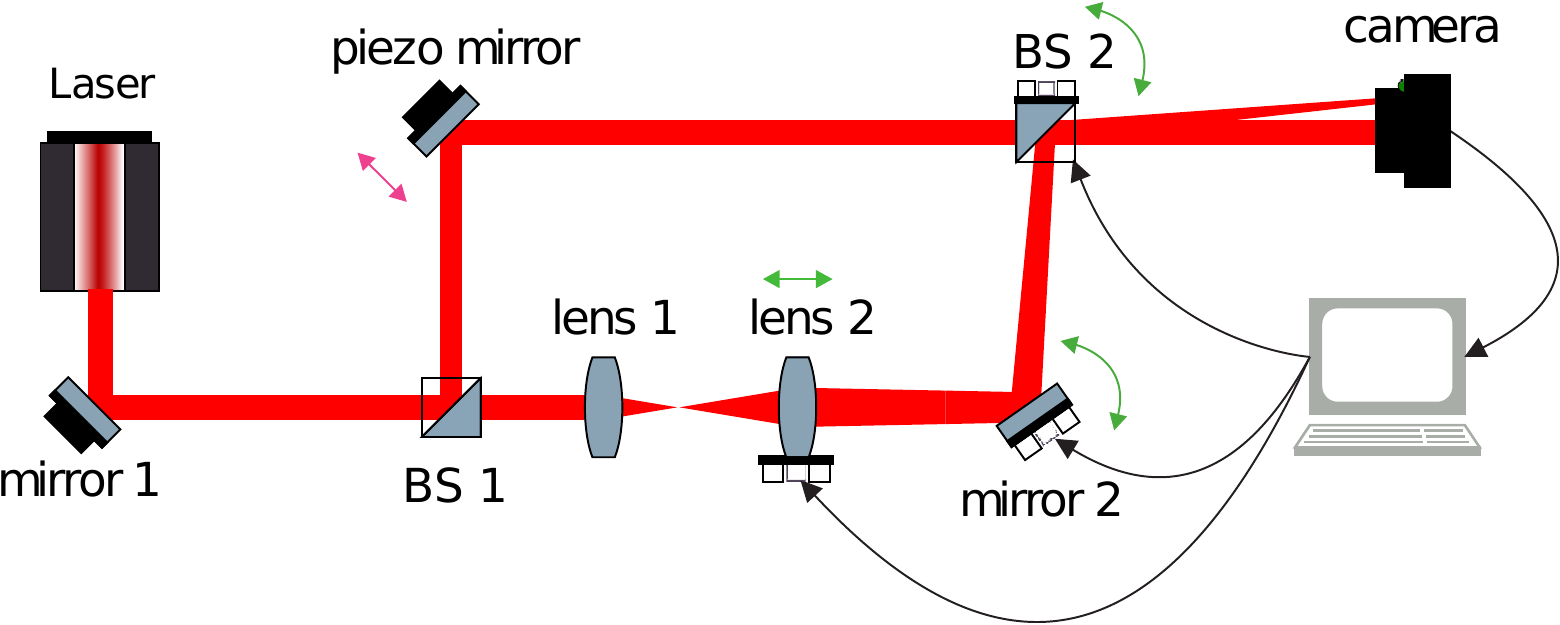}
  \centering
  \caption{Conceptual scheme of the Mach-Zehnder interferometer. Lens 2, mirror 2 and BS 2 are motorized optical elements controlled by an RL agent.}
  \label{fig:scheme}
\end{figure}

Interference is a physical phenomenon that results from coherent addition of amplitudes of two or more overlapping waves; the resulting amplitude depends on the relative phase of the component waves. Interferometers, which are among the main instruments of experimental optics, use interference to precisely measure this phase difference. 

In this paper we consider an MZI displayed in Fig.~\ref{fig:scheme}. A collimated (parallel) laser beam is divided by a beam splitter (BS 1). The two resulting beams propagate through different paths before being recombined, with the help of steering mirrors, by another beam splitter (BS 2) and viewed by a camera. One of the mirrors is mounted on a piezoelectric transducer to vary the relative phase $\varphi$ of the beams. One of the beam paths (the lower one in Fig.~\ref{fig:scheme}) contains a system of two lenses (telescope), which control the beam divergence and size. We use motorized mounts for mirror 2 and BS 2; lens 2 is mounted on a motorized translation stage. Aligning the interferometer requires precise matching of the two beams in terms of their transverse positions, directions, radii and divergences.

\begin{figure}
  \includegraphics[width=0.86\linewidth]{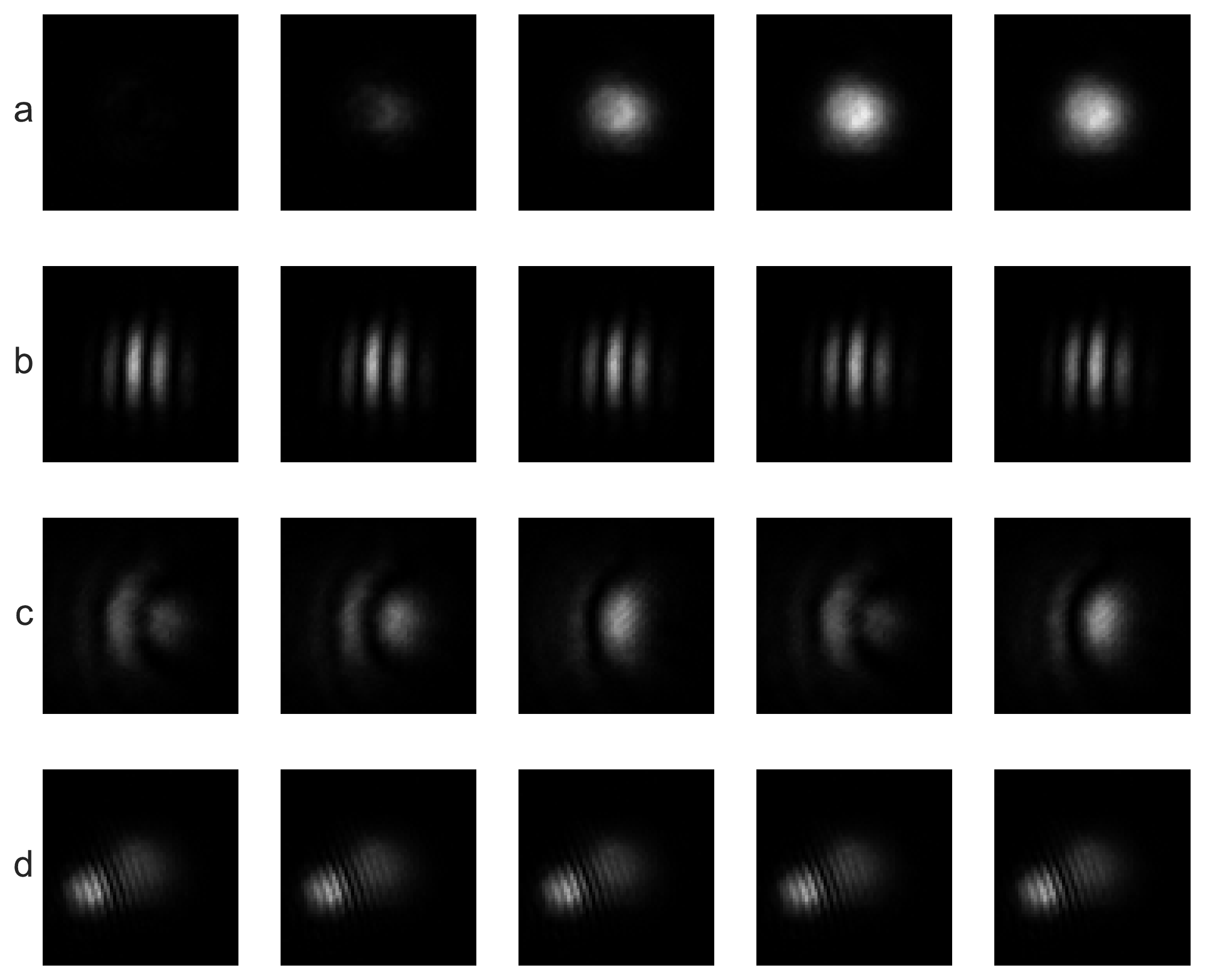}
  \centering
  \caption{Camera images from the Mach-Zehnder interferometer. (b-d) Example images acquired with a misaligned interferometer. The images in each row correspond to varying path length differences of the interferometer arms.}
  \label{fig:patterns}
\end{figure}

The images from the camera serve as observations for our RL agent. Examples of acquired images are shown in Fig.~\ref{fig:patterns}. The curved shape of the interference fringes is a consequence of different divergences of the two beams. We apply an asymmetric sawtooth voltage pattern to the piezo, akin to Ref.~\citep{sorokin2020interferobot}, to view the temporal dynamics of fringes. The amplitude of the piezo motion corresponds to a phase difference of about $2\pi$. For a misaligned interferometer, the piezo motion causes transverse displacement of the fringes visible in Fig.~\ref{fig:patterns} (b-d), which permit the agent to extract the information about the sign of the difference of the beams' angles. When fully aligned, the interference will appear as a single blinking spot [Fig.~\ref{fig:patterns} (a)].

The complexity of alignment arises because changing the angular orientations of the mirrors
%mirror's rotation 
simultaneously affects the position and angle of the lower beam, while the lens movement changes the lower beam radius and divergence. Moreover, images received from the camera can be corrupted by noise, aberrations, ambient light and dust.
%flares, and motes. 
The problem is additionally complicated by the inaccuracy of mounts and positioners
%of optomechanical mirrors, as well as the movement of the lens,
whose relative action noise is about $4\%$.

The position-dependent intensity of the interference pattern can be written as $I(x,y,\varphi) = \frac12|E_{\rm upper}(x,y)e^{i\varphi} + E_{\rm lower}(x,y)|^{2}$, where $E_{\rm upper/lower}(x,y)$ is the amplitude of each beam. The quality metric of alignment, \emph{visibility}, is defined as

\begin{equation}
    V = \dfrac{\max_{\varphi}(I_{\rm tot}(\varphi)) - \min_{\varphi}(I_{\rm tot}(\varphi))}{\max_{\varphi}(I_{\rm tot}(\varphi)) + \min_{\varphi}(I_{\rm tot}(\varphi))}
    \label{eq:visib_def}
\end{equation}

where $I_{\rm tot}(\varphi) = \iint I(x,y,\varphi)dxdy$ is a total light intensity in the image plane with the phase difference $\varphi$ between the arms. The intensity depends on time because of periodic piezo movement that varies the phase difference between the two interferometer arms. The visibility lies in $[0, 1]$ range, where $V=1$ corresponds to a perfectly aligned interferometer. 

%In the simulation, we approximate $S$-plane with $64\times64$ pixels image, and continuous piezo mirror period time with 16 equidistant discrete time steps. ???

%===============================================================================
%%defined by tuple $(\mathcal{S}, \mathcal{A}, \mathcal{O}, F, U, R)$
\section{Background}
We consider a standard partially observable Markov decision process (POMDP) setting where the purpose of an agent is to maximize its cumulative reward during the policy execution. The observation of the environment at timestep $t$ is defined as $o_t \in \mathcal{O}$ and is sampled from the  distribution $o_t \sim U(o_t|s_t)$, where $s_t \in \mathcal{S}$ is the state. The policy of our agent is a  deterministic function $a_t=\pi(o_t)$ yielding an action, which is an element of the action space $\mathcal{A}$. Following the action, the agent receives a reward $r_t: \mathcal{S} \times \mathcal{A} \to \mathcal{R}$ and the next observation $o_{t+1} \sim U(o_{t+1}|s_{t+1})$, where $s_{t+1}$ is produced according to a latent transition distribution $s_{t+1} \sim F(s_{t+1}|s_t, a_t)$. The discounted reward sum (return) from timestep $t$ in each episode is $R_t = \sum^{T}_{i=t}\gamma^{i-t}r_i$, where $\gamma \in [0,1]$ is the discount factor and $T$ is the horizon of the episode.

The agent's objective is to learn the policy $\pi^*$ that maximizes the expected return $J(\pi) = \mathbb{E}_{\tau\sim p(\tau | \pi)}[R_0]$ where $p(\tau | \pi)$ is the distribution of trajectories $\tau = (o_0, a_0, o_1, a_1, ..., a_{T-1}, o_{T})$ produced by policy $\pi$:
\begin{equation}
    \label{eq:optEq}
    \pi^* = \argmax_{\pi}J(\pi).
\end{equation}

\textbf{Policy gradient. }
The policy $\pi$ is defined by parameters $\theta$ (i.e.~$\pi=\pi_{\theta}$), so the optimization \eqref{eq:optEq} is performed with respect to the parameters: $\theta^* = \argmax_{\theta}J(\pi_{\theta})$. A common approach to such a problem is policy gradient \citep{sutton1999policy}, which iteratively improves the policy in terms of expected return via gradient ascent:
\begin{equation}
    \theta_{t+1} = \theta_{t} + \alpha  \frac{\partial J(\pi_{\theta})}{\partial \theta},
\end{equation}
where $\alpha$ is the learning rate. The algorithms of this family vary by the method of approximating the unknown $J(\pi_\theta)$.

A common approach to continuous-action Markov decision processes is TD3 \citep{fujimoto2018addressing} which is an extension of another popular algorithm DDPG \citep{lillicrap2015continuous}. This algorithm uses three neural networks: one (actor) for a deterministic policy and two (critics) for evaluating the action-state values $Q(s_t,a_t)$. Using two critics allows the algorithm to suffer less from overestimating the Q-values; in addition, the algorithm smoothens the Q-functions by adding Gaussian noise to the target actions when updating the parameters of the critics. Gaussian noise is also added to each policy action during the training to encourage exploration.
% Данный алгоритм использует нейронную сеть, определенную набором параметров $\theta$, для реализации детерминированной политики $a = \pi_{\theta}(o)$. Для обучения такой политики  функция $J(\pi_\theta)$ приближается с помощью двух Q-функций, заданных нейронными сетями что уменьшает переоценку Q-значений. 

% Так TD3 использует две Q-функции вместо одной, для предотвращения переоценки $J(\pi_\theta)$, реже обновляет политику, а также сглаживает Q-функции, добавляя шум к целевому действию. 
% Добавить про TD3. Exploration noise ???

\textbf{Domain randomization. }
%Frequently it is infeasible to train an agent in a real environment. The main reasons for that is high time consuming - the majority of RL algorithms need about $10^6$ steps of interaction with the environment, and unsafeness of execution transitional policies during training. One of the approaches to overcome such restrictions and successfully apply RL to real-world tasks is domain randomization. The method allows to train the algorithm in the precise simulation of the real environment and then transfer acquired policy. 
To lower the performance losses associated with the transferring of the agent from a simulated to real-world environment, domain randomization is used. The agent is trained  for a set of tasks with different dynamics $\hat{F}(s_t|s_{t-1}, a_{t-1}, \mu_{\rm sim}) \approx F(s_t|s_{t-1}, a_{t-1}, \mu_{\rm real})$ and sensor noise models $\hat{U}(o_t|s_t, \mu_{\rm sim}) \approx U(o_t|s_t, \mu_{\rm real})$ where $\mu_{\rm sim/real}$ are sets of environment parameters. Whereas we do not know the actual $\mu_{\rm real}$ of the real environment, varying $\mu_{\rm sim}$ in a range containing $\mu_{\rm real}$ helps to improve generalization and transfer quality.

% Frequently it is infeasible to train an agent in a real environment. The main reason for that is high time consuming - the majority of RL algorithms need about $10^6$ steps of interaction with the environment, and unsafeness of execution transitional policies during training. One of the approaches to overcome such restrictions and successfully apply RL to real-world tasks is domain randomization. The method allows to train the algorithm in the precise simulation of the real environment and then transfer acquired policy. To lower the performance losses after transfer the method proposes to train agent for set of tasks, having different dynamics $\hat{p}(o_t|o_{t-1}, a_{t-1}, \mu_{sim}) \approx p(o_t|o_{t-1}, a_{t-1}, \mu_{real})$ where $\mu$ is a set of environment parameters, i.e. friction for a pushing task\citep{peng2018sim}. Whereas we do not know the actual $\mu_{real}$ of the real environment, varying $\mu_{sim}$ in a range containing $\mu_{real}$ helps to improve generalization and transfer quality.

%===============================================================================

\section{Our method}

In this work, we apply continuous control reinforcement learning methods to align the MZI. As mentioned above, the alignment is performed step by step and the actions depend on the interference patterns observed by the agent, so it can be naturally viewed as a POMDP. 

The agent is trained in simulation and evaluated on a real interferometer. The training algorithm is listed in Appendix A and the MZI simulator is described in Appendix B.

%In contrast with conventional function optimization methods here we need to analyze interference images to predict the next action. %%Also noisy actions add uncertainty in the current state. 

%%Our purpose is to train a continuous action policy that will outperform the human-expert and baseline DQN approach on the complicated task of alignment MZI in terms of alignment speed and quality.

\begin{table}[b]
    \centering
    \begin{tabular}{|c|c|c|c|c|c|} 
     \hline
     Element angle / position & mirror 2, x & mirror 2, y & BS 2, x & BS 2, y & lens 2 \\
     \hline
     Amplitude of max deflection & $2.6 \cdot 10^{-3}$ & $1.8 \cdot 10^{-3}$ & $1.3 \cdot 10^{-3}$ & $0.9 \cdot 10^{-3}$ & $7.5$ \\
     \hline
     
    \end{tabular}
    \vskip 0.05in
    \caption{Maximum deflection of each optical element. Mirrors angles are given in radians, lens positions in millimeters.}
    \label{table:0}
\end{table}

\textbf{State, observation and action.}
% The observation is the set of 16 consecutive 64 × 64 images acquired by the high-frequency camera during one period of piezo. Actions are 5D vectors specifying deflections of both mirrors along $x$ and $y$ axis and relative displacement of the lens. Actions represent relative offsets from current mirrors and lens positions. However, optomechanical mirrors and stage support actions with amplitudes in the $[1e^{-6}, 1]$ range. We restrict action amplitudes to $[2.5e^{-3}, 1]$ interval because lower actions do not produce an observable change in camera images.
The state of the environment is a vector consisting of the transverse position $(x,y)$, direction angle $(\alpha_x,\alpha_y)$, radius $r_{\rm lower}$ and divergence curvature $\rho$ of the lower beam in the plane of the camera. The position and transverse direction of the upper beam are assumed to be zero. This state is fully determined by the angles of the mirror and the beam splitter and the position of the lens. The range of allowed states, listed in Table \ref{table:0}, is  restricted by the requirement that the beam remains visible on the camera and the position of the lens stays within the travel range of the translation stage.

The observation is a set of 16 consecutive $64\times64$ images acquired by the camera during one period of the piezo mirror. %The mirrors and lens positions lie in the $[-1,1]$ range, where $1$ correspond to maximum deflection from aligned state such that both beams are observable on the camera. 
The actions are five-dimensional vectors specifying the relative angular deflections of both mirrors along the $x$ and $y$ axes and the linear displacement of the lens with respect to their current positions. Each action vector component lies in the $[-1,1]$ range, where the values of $\pm 1$ correspond to the maximum and minimum values as listed in Table \ref{table:0}. %  therefore if applying action overflows the acceptable position range it is clipped either to $1$ or $-1$.
The absolute values of each action are restricted to the interval $[2.5 \cdot 10^{-3}, 1]$, because smaller actions do not produce observable changes in the interference pattern and fall within the uncertainty range of motorized mounts.

\textbf{Episode and reset.}  
% $\alpha_x, \alpha_y, \beta_x, \beta_y, lens_{pos}$ are sampled from corresponding intervals: $[-\alpha^{max}_x,\alpha^{max}_x], [-\alpha^{max}_y,\alpha^{max}_y], [-\beta^{max}_x,\beta^{max}_x], [-\beta^{max}_y,\beta^{max}_y], [-lens_{pos}^{max}, lens_{pos}^{max}]$
% All our experiments were carried out at MZI which scheme is presented in Figure \ref{fig:scheme} and its simulation. The episode length is set to 100 actions, which is sufficient for both agents to align the interferometer. At the beginning of each episode, the mirrors angles and lens position are set to random values within their corresponding maximum values, the values guarantee that both beams are observable on camera. Each action is clipped to stay in such an area. Visibility is the metric of interferometer alignment but it exponentially decreases with positions and k-vectors as per Equation (5), being sparse. Therefore the reward is $R = V - log(1-V)$ where $V$ is visibility consists of two terms: the first one encourages the linear increase of visibility and the second one logarithmic increase, making $V = 0.99$ much higher rewarded than $V=0.98$. We also introduce low negative reward $R = -0.04$ and episode termination for overflowing actions, that are clipped in simulation but can lead to unsafe behaviour in the real setting.

To compare the performance of our agent with that of \citet{sorokin2020interferobot}, we keep the episode length equal to 100 actions. At the beginning of each episode, the beam in the lower path is misaligned by setting the mirror angles and lens position to random values within the allowed range.

\textbf{Reward. }
Under the normal alignment procedure, the  reward at each step is positive and consists of two terms:
\begin{equation}\label{eq:reward}
  R =      V - \log(1-V) .    
\end{equation}
   
The first term is the visibility, which, as discussed above, is the primary metric for the interferometer alignment quality. 
%However, its direct use as reward in RL algorithms is not optimal because it exponentially decreases with positions and k-vectors as per Equation (5), being sparse. Therefore we use reward:
The second term rewards high quality of the final alignment, which is critical for optical experiments; it tends to infinity for $V\to1$ \cite{sorokin2020interferobot}. 

However, if the agent proposes an action that takes one of the controls out of the boundaries defined in Table \ref{table:0}, the episode is terminated to avoid damage to the equipment. Additionally, the agent is penalized with the reward of $P=-0.04$. This penalty is important during early stages  of the training, when  the reward \eqref{eq:reward} is normally close to zero. A small negative reward for unsafe actions will train the agent to be aware of the bounds, but will not discourage exploration. The specific value of $P$ was handpicked via experimentation. On the other hand, when the agent is well-trained, it  normally receives a significant positive reward for each step. At this stage, terminating an episode has a major negative effect on the return, thereby strongly discouraging unsafe actions.

%Finally, $R_{\rm penalize} =$  is a low negative reward for overflowing actions. We also terminate episode in case of overflowing actions to punish the agent for unsafe behaviour in the real environment. 
\begin{figure}
  \includegraphics[width=1\linewidth]{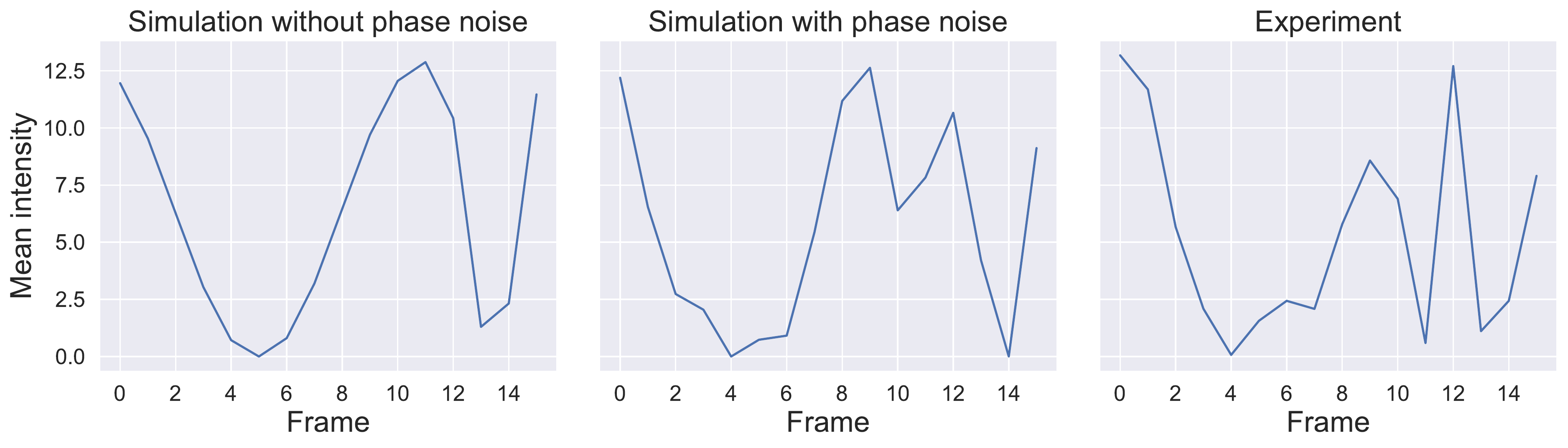}
  \centering
  \caption{Effect of phase noise. The plots display the integrated intensity of the interference pattern as the path length difference is varied. The behavior becomes closer to real after applying phase noise.}
  \label{fig:piezo_noise}
\end{figure}

\textbf{Domain randomization. }
To learn a more generalized policy and facilitate sim-to-real transfer, we introduce several domain randomizations, which are used during the training. The only randomization that changes the dynamic of the environment $F(s_{t+1}|s_{t},a_{t})$ is the beam radius randomization within $\pm20\%$ of the measured radius $r=0.71$ mm. The randomization is applied at the beginning of each episode. 

Additionally, we introduce the following randomizations of the observation $U(o_t|s_t)$. First, to address the noise of the camera detector, scattering, air fluctuations and dust effects, we add Gaussian noise to each pixel acquired by the camera. The standard deviation of the noise is 20\% with respect to the simulated intensity value. %The other important source of discrepancy is the uncertainty in the camera sensor  therefore the following randomizations are applied on every step of environment execution. First, for every $(x,y)$ pair we sample $E_{\rm upper}(x, y)$ and $E_{\rm lower}(x, y)$ from Gaussian distribution $\mathcal{N}(E_{\rm upper/lower}, 0.2)$ . 
Second, we rearrange cyclically the images within the video sequence to account for the randomness of the camera trigger. In addition, we randomize the duty cycle (fraction of time spent on the forward and backward passes) of the piezo. Third, we  vary the beams' brightness by $\pm30\%$ to model the variance of the camera exposure.
All these randomizations are applied to each time step (i.e.~they are constant for the 16 frames acquired in each step).

The randomizations described above have  been used in \citet{sorokin2020interferobot}  and have proven their effectiveness. An additional randomization we introduce in this work addresses irregularities in the motion of the piezo mirror and the frame rate, as well as fluctuations in the air density in different arms of the interferometer, all of which lead to random variation of the optical path length difference between the two MZI arms. To simulate these effects, we add Gaussian noise with the standard deviation of $0.5$ rad to this phase difference. %propose to use randomization that we name phase noise. For each frame $k$, it samples the phase between beams either from the distribution $\mathcal{N}(\frac{2\pi * k}{N_{\rm forw}}, 0.5)$ or from $\mathcal{N}(\frac{2\pi * k}{N_{\rm backw}}, 0.5)$ depending on whether it is a forward or a backward pass, $N_{\rm forw/backw}$ is a number of frames taken during the pass.

%  We also add relative action noise of $6\%$, but it is not affect training because of the exploration noise existing, however, it helps to choose the most effective model in the validation phase.
% Third, we simulate uneven piezo phase distance between images by adding Gaussian noise $n = \mathcal{N}(0, 0.5)$ to every equidistant piezo phase

\textbf{Action rescaling. } 
As the alignment progresses, the agent's actions become smaller and more precise. The typical action magnitude decreases during an episode by about two orders (as illustrated in the experimental Section \ref{sec:Exp} below). It is therefore desirable that the exploration noise also decrease with the action magnitude. We satisfy this by setting the agent's neural network to output a ``raw action" value $a_0\in[-1,1]$, from which the actual action is calculated according to %By using actions sampled from standard Gaussian distribution we either strongly eliminate the probability of choosing low-amplitudes actions, or train with extremely small exploration noise that leads to a lack of exploration and stack in local optima. To make different actions amplitudes $0.01, 0.1, 1$ have a similar probability of being produced by an agent we applied a special function:
\begin{equation}\label{eq:rescale}
a =
   \begin{cases}
    {\rm sign}(a_0) \cdot 1000^{|a_0| - 1}  & \quad \text{if $|a_0| > 0.17$} 
    \\
    0  & \quad \text{if $|a_0| \leq 0.17$}
  \end{cases}
\end{equation}
This transformation produces rescaled actions with absolute values $|a|\in\{0\}\cup[2.5 \cdot 10^{-3}, 1]$.  %range and $0$ for every action unobservable on camera images.
%\textcolor{red}{The exploration noise is first added to “raw action” and then action rescaler is applied, it is also significant that noisy “raw action” $a_0$  is stored in the replay buffer, while a is executed in the environment}

\textbf{Algorithm and network architecture. }
We use the standard TD3 algorithm with handpicked hyperparameters to produce a deterministic policy, which yields the raw action $a_0$ as described above. %$a_t = \pi(o_t)$, but $a_t$ is not directly sent to environment, and instead goes through action rescale $a^{env}_t = f(a_t)$. That allows us to train policy in an logarithmic space of actions.
For both critics and the actor, we use the  VGG-16 \citep{simonyan2014very} architecture, modified as follows. The number of convolutional layers in the encoder is set to 8, followed by three MLP layers, with no dropout. We have chosen this architecture because max pooling operations help reducing the overfitting and sensitivity to individual pixel noise. %making the output continuous action $a_0$ less sensitive to values of every particular feature %input pixel
%and concentrates only on the most important ones. 
We use orthogonal initialization  in all models as we found it to ensure faster convergence.

We train the agent with a discount factor $\gamma = 0.8$; this relatively short reward sight inspires our agent to  reach high visibility faster. A Gaussian exploration noise is added to the raw action $a_0$ with the standard deviation decreasing exponentially from 0.5 to 0.02 during the training. Although the variance of this noise is independent of the magnitude of $a_0$, its effect on the actual action $a$ is proportional to its magnitude due to the exponential dependence (\ref{eq:rescale}). The total number of steps is $10^6$ and the replay buffer size is $10^5$. Updates are performed every ten steps. The whole training on an NVidia RTX 2080 GPU takes about 26 hours.

% \textbf{Twin Delayed Deep Deterministic policy gradient (TD3)}

% One of the common approaches to continuous control tasks is TD3\citep{fujimoto2018addressing} which is the extension of another popular algorithm DDPG\citep{lillicrap2015continuous}. We use standard TD3 algorithm with handpicked hyperparameters to produce deterministic policy $a_t = \pi(o_t)$, but $a_t$ is not directly sent to environment, and instead goes through action rescale $a^{env}_t = f(a_t)$. That allows us to train policy in an exponential space of actions. We also add relative action noise of $6\%$, but it is not affect training because of the exploration noise existing, however, it helps to choose the most effective model in the validation phase. 

%===============================================================================
\section{Experimental evaluation}\label{sec:Exp}

\begin{figure}[ht]
    \begin{subfigure}[t]{0.5\textwidth}
        \includegraphics[width=\textwidth]{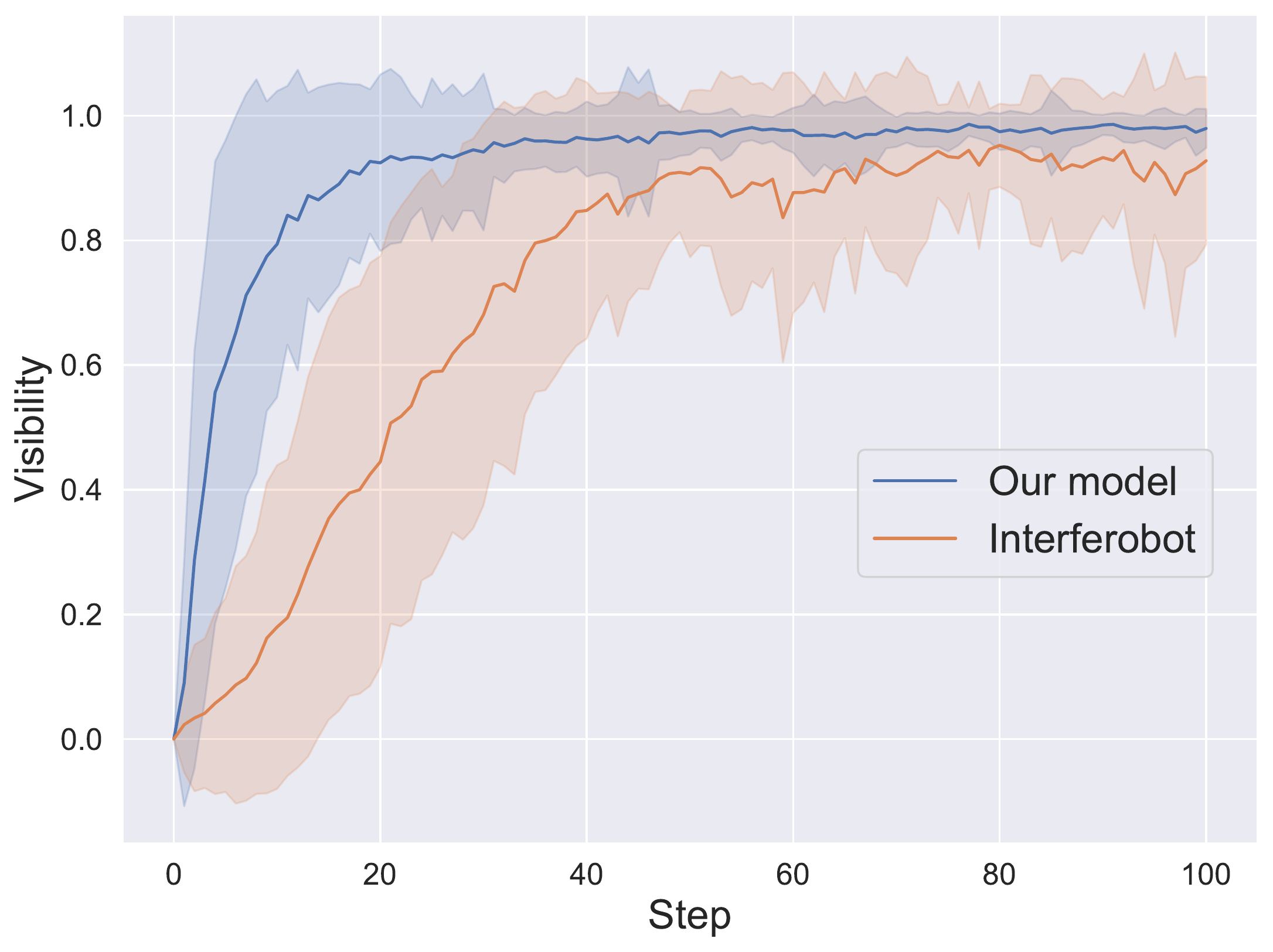}
        \caption{Mean visibility per step}
        \label{fig:results_a}
    \end{subfigure}
    \begin{subfigure}[t]{0.5\textwidth}
        \includegraphics[width=\textwidth]{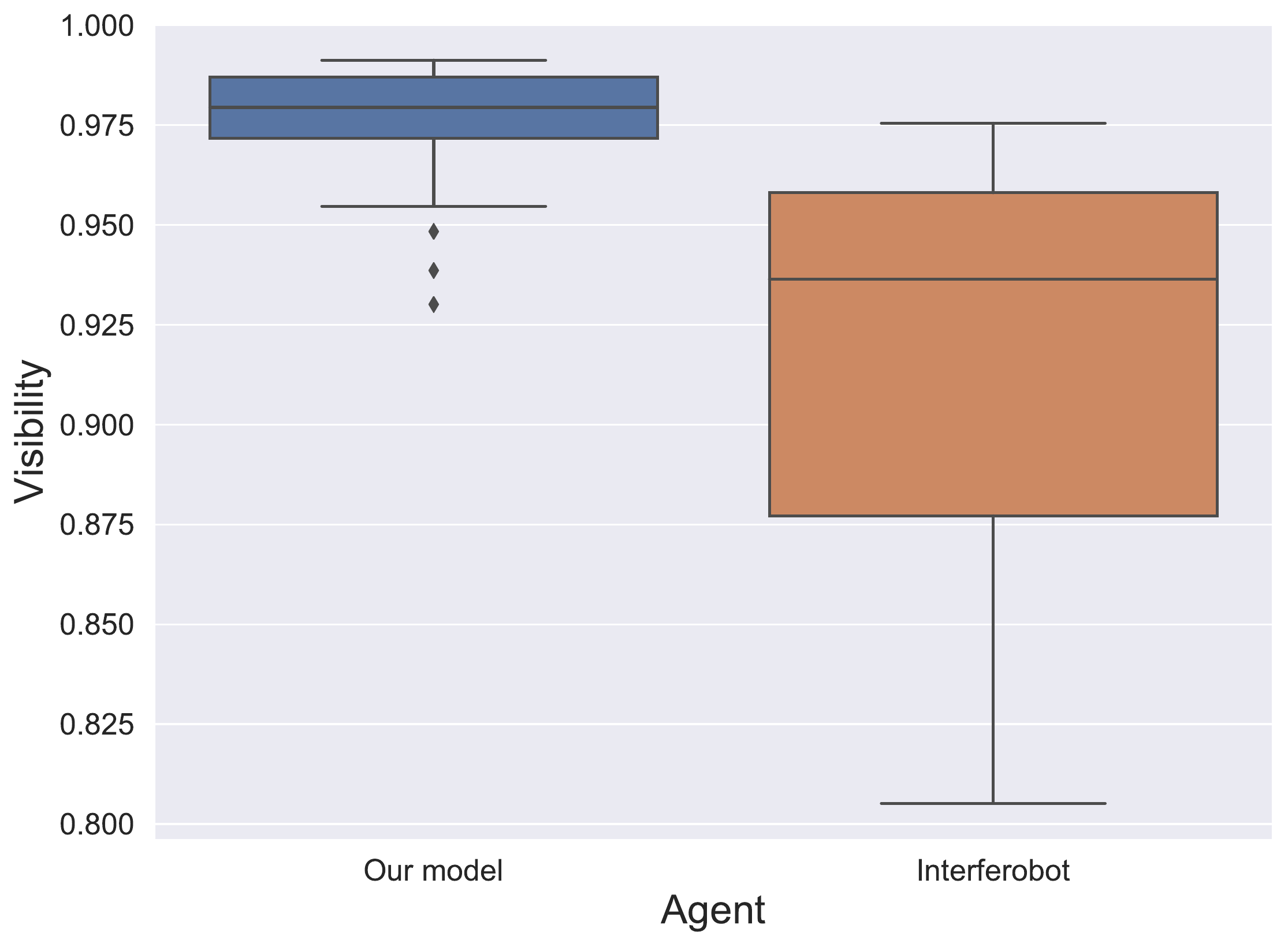}
        \caption{Box plot of visibility}
        \label{fig:results_b}
    \end{subfigure}
    \caption{Comparison of Interferobot and our agent, demonstrating the advantage of the latter in convergence speed and the final alignment quality. As evident from the box plot (b), the 25-75  interquartile range for our agent is significantly narrower  than that of Interferobot, indicating higher stability of the former. %a) Visibility as a function of step in an episode, averaged over 50 episodes. Fig. \ref{fig:results_b} shows that the interquartile range for the TD3 agent is significantly lower then for previous DQN agent indicating that it is more stable. 
    All results are averaged over 50 episodes.}
    \label{fig:results}
\end{figure}

For our experiments, we build an MZI according to the scheme shown in Fig.~\ref{fig:scheme}. We use a continuous HeNe laser with wavelength $\lambda = 632$ nm, Newport Picomotor mirror mounts, a Standa 8MT167-25LS linear translation stage, a CMOS camera with 16 fps acquisition rate, and a high-bandwidth photodetector to precisely measure the visibility. In addition to our agent, we also train and evaluate the original Interferobot \cite{sorokin2020interferobot} with the action space extended to include the lens movement. For each agent, we run $50$ episodes of evaluation with $100$ timesteps.  Figure \ref{fig:results} shows the evaluation results. Our agent significantly outperforms Interferobot in terms of the alignment time, final visibility (averaged over the last 40 steps of each episode) and its variance. 

%As the original  demonstrated an advantage over human-expert in previous work, our new model can be considered superhuman.
Additionally, our agent has been compared with a human expert, who  aligned the interferometer 10 times by manually turning the controls on the physical setup. As evidenced by Table \ref{table:human}, our agent surpasses the human both in the speed and quality of the alignment. 

\begin{table}
\centering
    \begin{tabular}{lccc}\toprule
        &V $\ge 0.92$ & V $\ge 0.95$ & V $\ge 0.98$\\\midrule
        Human &  93.9 (\textbf{0\%})  & 103.6 (\textbf{0\%}) & 129.6 (10\%)\\
        TD3 (our agent) &  \textbf{56.16} (\textbf{0\%}) & \textbf{75.06} (\textbf{0\%}) & \textbf{120.1} (\textbf{4\%})\\
        Interferobot &  98.7 (7.6\%) & 116.1 (7.6\%) & 156.4 (10.6\%)\\
        \bottomrule
    \end{tabular}
    \vskip 0.05in
    \caption{Comparison with a human expert and the original Interferobot. The time required to reach the visibility thresholds of 0.92, 0.95 and 0.98 are shown, together with the percentage of episodes the threshold has not been reached (in parentheses).}\label{Tab1}
    \label{table:human}
\end{table}

\begin{figure}[ht]
    \begin{subfigure}[t]{0.5\textwidth}
        \includegraphics[width=\textwidth]{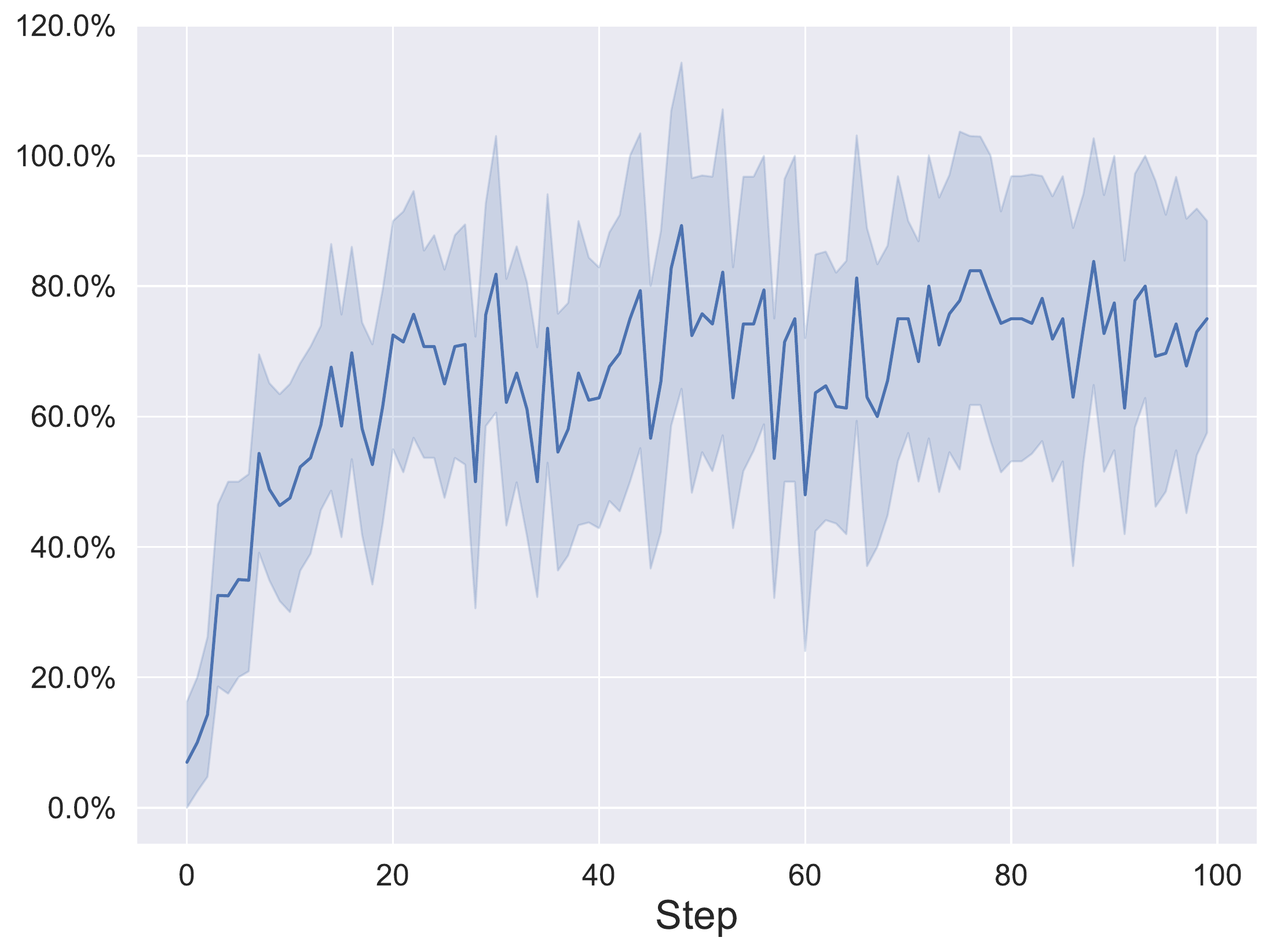}
        \caption{Percentage of parallel actions.}
        \label{fig:parallel_actions}
    \end{subfigure}
    \quad
    \begin{subfigure}[t]{0.5\textwidth}
        \includegraphics[width=\textwidth]{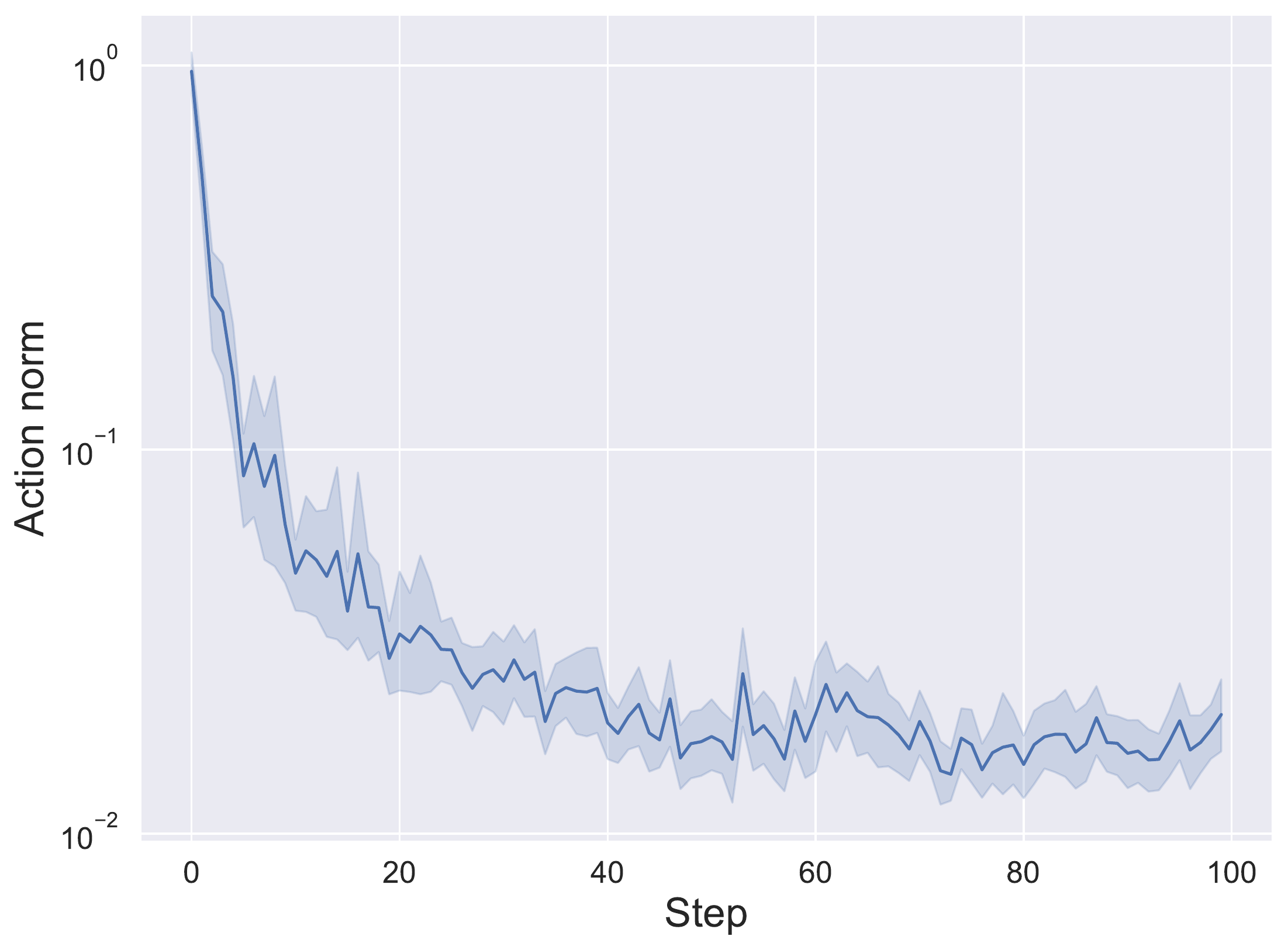}
        \caption{Norm of actions.}
        \label{fig:act_norm}
    \end{subfigure}
    \caption{Interpreting the agent's policy. a) Percentage of actions that change the distance between beams while conserving the directions increases with the step number. b) Mean norm of actions decreases with the step number.}
     \label{fig:parallel_norm}
\end{figure}

%To align an interferometer our agent needs to precisely superpose two laser beams both in position and direction. This process is complicated by the fact that each of the two mirrors changes both position and direction of the beam. 
A common trick used by human experts in aligning optics is to turn two mirrors steering the same beam by equal and opposite angles. This action moves the position of the beam without changing its direction. Fig.~\ref{fig:parallel_actions} shows that our agent learns this method and uses it extensively: by the end of the procedure, about 70\% of the actions contain such parallel movements. 
%the two beams good strategy, in this case, would be to align both beams parallel at the first stage and move the second beam along the camera plane parallel to the first beam. This can be achieved when both mirrors are rotated at the same angle. It is seen that at the first 10 steps this percentage is relatively small but after it grows up to 80\% meaning that our agent has learned such actions by itself without any prior knowledge and special constraints and use them at the final steps for fine-tuning the interferometer. 
Fig. \ref{fig:act_norm} shows the action norm as a function of the step number. The descending trend means that at the beginning our agent uses large actions to obtain rough alignment and decreases the action size to fine-tune the interferometer.

Table \ref{table:1} presents the results of an ablation study. A standard TD3 agent does not outperform the original Interferobot and achieves a mean visibility of $V = 0.83$. Action rescaling and phase noise randomizations significantly boost the performance of our agent, leading to a visibility of $V = 0.98$. It is notable that the phase noise randomization slightly decreases the mean visibility of the Interferobot but improves the standard deviation. This bias-variance trade-off can be due to the relative simplicity of the Interferobot model which cannot capture richer observations produced by this randomization.  

\begin{table}[ht]
\centering
\begin{tabular}{|l |c| c|} 
 \hline
 Model & mean visibility for last 40 step & standard deviation \\
 \hline
 TD3 + AR + PN & \textbf{0.98} & \textbf{0.03} \\
\hline
TD3 + AR & 0.95 & 0.06\\
\hline
Interferobot  & 0.92 & 0.12\\
\hline
Interferobot + PN & 0.90 & 0.08\\
\hline
TD3& 0.83 & 0.18\\
\hline
\end{tabular}
\vskip 0.05in
\caption{Comparative evaluation with ablated agents and the original Interferobot \cite{sorokin2020interferobot}. PN: phase noise; AR: action rescaling.}
\label{table:1}
\end{table}

%===============================================================================
\section{Conclusion}

We demonstrated a novel RL algorithm for the automatic alignment of an optical interferometer, which contains a system of lenses in one of its optical paths. Such a setup results in richer observation and action spaces and constiututes a better approximation of a general problem of achieving mode matching between two arbitrary Gaussian beams.  Our reinforcement learning agent with a continuous action space solves the alignment problem successfully, surpassing both a previous solution and a human expert. Important innovations include exponential scaling of the action space, an additional domain randomization that helps the agent generalize to a real setup after training in simulation as well as a discrete reward structure that trains the agent to avoid unsafe actions. Our solution will work for any Mach-Zehnder interferometer with $2\times2$ degrees of freedom (mirror angles) controlling the  position and direction, and one additional degree of freedom (lens position) controlling the width and divergence of one of the beams. %We demonstrate a significant advantage of our agent over the previous discrete one in the experiments on the real interferometer.
%We also investigated the impact of each improvement on the agent's performance and examined the agent's strategy against a known interferometer tuning algorithm.

% which changed the divergence and radius of the beam; this significantly complicated the task, making the observed interference patterns less interpretable

% We have shown that the existing approach to the alignment of Mach Zehnder Interferometer can be extended to a more complex optical system including lens and that it is possible to speed up the alignment of Mach Zehnder Interferometer, without quality loss. We proposed additional randomizations and the way to leverage actions with different orders of amplitudes, showing their fruit fullness. 

\section{Acknowledgments}
We acknowledge support from Russian Science Foundation (19-71-10092).

%===============================================================================

% The maximum paper length is 8 pages excluding references and acknowledgements, and 10 pages including references and acknowledgements

%\clearpage
% The acknowledgments are automatically included only in the final and preprint versions of the paper.
%\acknowledgments{If a paper is accepted, the final camera-ready version will (and probably should) include acknowledgments. All acknowledgments go at the end of the paper, including thanks to reviewers who gave useful comments, to colleagues who contributed to the ideas, and to funding agencies and corporate sponsors that provided financial support.}

%===============================================================================

\bibliography{article}  % .bib
\newpage
\section*{Appendix}

\subsection*{A. The pseudo code and hyperparameters}
        % \scalebox{0.9}{
        % \begin{minipage}[!t]{1\textwidth}
        \begin{algorithm}[H]
        \caption{TD3 with action rescaling}
        \label{alg1}
        \begin{algorithmic}[1]
        \STATE Input: initial policy parameters $\theta$, Q-function parameters $\phi_1$, $\phi_2$, empty replay buffer $\mathcal{D}$
        \STATE Set target parameters equal to main parameters $\theta_{\text{targ}} \leftarrow \theta$, $\phi_{\text{targ},1} \leftarrow \phi_1$, $\phi_{\text{targ},2} \leftarrow \phi_2$
        \FOR{$i$ in range(number of steps)}
            \STATE Get observation $o$ and select action $a_0 = \text{clip}(\pi_{\theta}(o) + \epsilon, -1, 1)$, where $\epsilon \sim \mathcal{N}(0, \sigma_{\rm explore})$
            \STATE Compute rescaled action $a(a_0)$ according to Eq.~(\ref{eq:rescale}) \label{aсtRescale}
            \STATE Execute $a$ in the environment
            \STATE Get next observation $o'$, reward $r$, and done signal $d$ to indicate whether $o'$ is terminal
            \STATE Store $(o,a_0,r,o',d)$ in replay buffer $\mathcal{D}$ \label{aсtStore}
            \STATE If $o'$ is terminal, reset environment state.
            \IF{$i > \texttt{start\_train\_step}$}
                \FOR{$j$ in range(\texttt{num\_epochs})}
                    \STATE Randomly sample a batch of transitions, $B = \{ (o,a_0,r,o',d) \}$ from $\mathcal{D}$
                    \STATE Compute target actions
                    \begin{equation*}
                        a'(o') = \text{clip}\left(\pi_{\theta_{\text{targ}}}(o') + \text{clip}(\epsilon,-c,c), -1, 1\right), \;\;\;\;\; \epsilon \sim \mathcal{N}(0, \sigma_{\rm targ})
                    \end{equation*}
                    \STATE Compute targets
                    \begin{equation*}
                        y(r,o',d) = r + \gamma (1-d) \min_{i=1,2} Q_{\phi_{\text{targ},i}}(o', a'(o'))
                    \end{equation*}
                    \STATE Update Q-functions by one step of gradient descent using
                    \begin{align*}
                        & \nabla_{\phi_i} \frac{1}{|B|}\sum_{(o,a_0,r,o',d) \in B} \left( Q_{\phi_i}(o,a_0) - y(r,o',d) \right)^2 & \text{for } i=1,2
                    \end{align*}
                    \IF{ $j \mod$ \texttt{policy\_delay} $ = 0$}
                        \STATE Update policy by one step of gradient ascent using
                        \begin{equation*}
                            \nabla_{\theta} \frac{1}{|B|}\sum_{o \in B}Q_{\phi_1}(o, \pi_{\theta}(o))
                        \end{equation*}
                        \STATE Update target networks with
                        \begin{align*}
                            \phi_{\text{targ},i} &\leftarrow \rho_{\rm polyak} \phi_{\text{targ}, i} + (1-\rho_{\rm polyak}) \phi_i & \text{for } i=1,2\\
                            \theta_{\text{targ}} &\leftarrow \rho_{\rm polyak} \theta_{\text{targ}} + (1-\rho_{\rm polyak}) \theta
                        \end{align*}
                    \ENDIF
                \ENDFOR
            \ENDIF
        \ENDFOR
        \end{algorithmic}
        \end{algorithm}
        % \end{minipage}%
        % }
        
        % \newpage
        \begin{table}[!ht]
        \begin{minipage}{.5\linewidth}
        \centering
        \begin{tabular}{|c|c||c|c|} 
        \hline
        batch\_size & 32 & $\pi$ lr & $10^{-5}$ \\
        \hline
        num\_epochs & 10 & $Q$ lr & $10^{-4}$ \\
        \hline
        policy\_delay & 1 & $\sigma_{\rm targ}$ & 0.2 \\
        \hline
        max\_grad\_norm & 10 & $c$ & 0.5 \\
        \hline
        start\_train\_step & $10^{4}$ steps & optimizer & Adam \\
        \hline
        $\rho_{\rm polyak}$ & 0.995 && \\
        \hline
        \end{tabular}
        \vskip 0.05in
        \caption{Training hyperparameters}
        \end{minipage}%
        \begin{minipage}{.5\linewidth}
        \centering
        \begin{tabular}{|c|c|} 
        \hline
        & Distance, mm \\
        \hline
        BS 1 $\to$ Mirror 2 & 300\\
        \hline
        Mirror 2 $\to$ BS 2 & 200\\
        \hline
        BS 2 $\to$ Camera & 100\\
        \hline
        BS 1 $\to$ Lens 1 & 50\\
        \hline
        $f_{\rm lens 1}=f_{\rm lens 2}$ & 50 \\
        \hline
        \end{tabular}
        \vskip 0.05in
        \caption{Setup parameters}
        \end{minipage}
        \end{table}

        In this section we show a full listing of our algorithm as well as the hyperparameters (aside from those mentioned in the main text) and the parameters of the physical setup.
        
        The exponential action scaling described in the article is performed in step~\ref{aсtRescale} of Algorithm~\ref{alg1}. Importantly, while the action $a$ is executed in the environment, the raw action $a_0$ is stored in the replay buffer $\mathcal{D}$ (step~\ref{aсtStore}).

% Можно совсем удалить аппендикс добавив фразу про ABCD формализм

\subsection*{B. The simulator}%To train our agent we need to model the Mach-Zehnder interferometer with a telescope. To do that we modify the simulator presented in \citep{sorokin2020interferobot}. 
The laser beams in the upper and lower arms of the interferometer are modelled with a Gaussian transverse profile. Their electric field amplitudes at a point in space with the coordinates $(x,y,z)$ are given by 
\begin{subequations}\label{beams}
\begin{equation}
    E_u={\rm Re}\left[\exp \left(-\frac{x^{2}+y^{2}}{r_u^{2}(z)}\right) \exp \left(-i\left(k_{z} z+ k\frac{x^2+y^2}{2\rho^2_u(z)} + \phi_{\rm piezo}(t)\right)\right)\right]
    \label{eq:upper_beam}
\end{equation}
\begin{equation}
    \label{eq:lowerr_beam}
    \begin{split}
        E_l={\rm Re}\left[\exp \left(-\frac{\left(x-x_{0}\right)^{2}+\left(y-y_{0}\right)^{2}}{r_l^{2}(z)}\right)  
        \exp \left(-i\left(k_{x} x+k_{y} y+k_{z} z + k\frac{x^2+y^2}{2\rho^2_l(z)} z\right)\right)\right]
    \end{split}
    \end{equation}
\end{subequations}

where the subscripts $u$ and $l$ respectively refer to the upper and lower beam, $(x_0, y_0)$ is the position of the centre of the lower beam [the upper beam is assumed centered at $(x,y)=(0,0)$], $z$ is the direction of beams propagation, %$\mathcal N$ the normalization factor, 
$r(z)$ the beam radius, $\rho(z)$ the wavefront curvature radius, $(k_x,k_y,k_z)$ the wave vector with $k=\sqrt{k_x^2+k_y^2+k_z^2}=2\pi/\lambda$, and $\phi_{\rm piezo}(t)$ the phase shift that produced by the piezo mirror periodic movement. We work in the paraxial approximation, such that $k_z \gg k_x, k_y$. 

Prior to the first beam splitter, the two beams have identical parameters. The propagation of the Gaussian beams is analyzed in the framework of the ABCD formalism, where the beam is characterized by a complex parameter $\dfrac{1}{q} = \dfrac{1}{\rho} - \dfrac{i \lambda}{\pi r^2}$ and the transformations of the beam as it propagates through the setup are given by  $q'=\dfrac{A q+B}{C q+D}$, 
%$$\begin{bmatrix} q_2 \\ 1 \end{bmatrix} = k \begin{bmatrix} A & B \\ C & D \end{bmatrix} \begin{bmatrix}q_1 \\ 1 \end{bmatrix}, $$
where, e.g., $\begin{bmatrix} A & B \\ C & D \end{bmatrix}=\begin{bmatrix} 1 & d \\ 0 & 1 \end{bmatrix}$ for free propagation over a distance $d$ and $\begin{bmatrix} A & B \\ C & D \end{bmatrix}=\begin{bmatrix} 1 & 0 \\ -1/f & 1 \end{bmatrix}$ for a thin lens of focal length $f$. 

The transverse field profiles (\ref{beams}) are calculated for both beams, after which the interference pattern and the visibility are evaluated according to 
$$I(x,y,z,t)=|E_u(x,y,z)+E_l(x,y,z)|^2$$%with centers $\vec{x}_{01}$, $\vec{x}_{02}$, wave vectors $\vec{k}_1$, $\vec{k}_2$ in an arbitrary point with coordinate vector $\vec{x}$ is given by: 
%\begin{equation}
%    I(\vec{x}, t) = A_1^2 + A_2^2 + 2 A_1 A_2 \cos \left( \phi_1 - \phi_2 + \phi_{\mathrm{piezo}}(t) \right),
%    \label{eq:intens}
%\end{equation}
and 
$$ V = \frac{            \max_{t}(I_{\rm tot}) - \min_t(I_{\rm tot})
            } {
                \max_{t}(I_{\rm tot}) + \min_t(I_{\rm tot})
            },
    \label{eq:visib}
$$
where $I_{\rm tot}(t) = \iint_{-\infty}^{+\infty} I(x, y, t) {\rm d}x{\rm d}y$ is the intensity integrated over the beam area; the maximum and minimum are taken over the period of piezo movement.

\end{document}